\documentclass{article} % For LaTeX2e
\usepackage{iclr2022_conference,times}

\usepackage{amsmath,amsfonts,bm}

\def\eqref#1{equation~\ref{#1}}
\def\1{\bm{1}}

\DeclareMathAlphabet{\mathsfit}{\encodingdefault}{\sfdefault}{m}{sl}
\SetMathAlphabet{\mathsfit}{bold}{\encodingdefault}{\sfdefault}{bx}{n}

\def\gO{{\mathcal{O}}}

\def\gX{{\mathcal{X}}}

\def\gSD{{\mathcal{SD}}}
\def\gRL{{\mathcal{RL}}}

\usepackage{hyperref}
\usepackage{url}

\usepackage{booktabs}
\usepackage{graphicx}
\usepackage{enumitem}
\usepackage{wrapfig}
\usepackage{amssymb}

\title{\fontsize{15.75pt}{15.75pt}\selectfont Global Counterfactual Explanations:\\Investigations, Implementations and Improvements}

\author{Dan Ley, Saumitra Mishra, Daniele Magazzeni\\
J.P. Morgan AI Research, London, UK\\
\texttt{\{dan.ley, saumitra.mishra, daniele.magazzeni\}@jpmorgan.com}
}

\iclrfinalcopy % Uncomment for camera-ready version, but NOT for submission.
\begin{document}

\maketitle

\begin{abstract}
Counterfactual explanations have been widely studied in explainability, with a range of application dependent methods emerging in fairness, recourse and model understanding. However, the major shortcoming associated with these methods is their inability to provide explanations beyond the local or instance-level. While some works touch upon the notion of a global explanation, typically suggesting to aggregate masses of local explanations in the hope of ascertaining global properties, few provide frameworks that are either reliable or computationally tractable.
Meanwhile, practitioners are requesting more efficient and interactive explainability tools.
We take this opportunity to investigate existing global methods, with a focus on implementing and improving Actionable Recourse Summaries (AReS), the only known global counterfactual explanation framework for recourse.
\end{abstract}

\section{Introduction}

% Added by Saumitra
% First attempt - to be refined
% Apart from speed, do we make any more contributions?
% Yes, our "Then-Generation" method does well on continuous features (Bottom-Right Experiments Figure) - Dan

Counterfactual explanations (CEs) identify input perturbations that result in desired predictions from machine learning (ML) models~\citep{verma2020counterfactual, karimi2021survey, stepin2021survey}. A key benefit of these explanations is their ability to offer recourse to affected individuals in certain scenarios (e.g., automated credit decisioning). Recent years have witnessed a surge of research therein, with a focus on identifying desirable properties of CEs, developing the methods to model those properties and understanding the weaknesses and vulnerabilities of the proposed methods~\citep{barocas2019hidden, venkatasubramanian2020philosophical, pawelczyk2021carla, slack2021counterfactual}.

Importantly, the research efforts so far have largely centred around local analysis, generating explanations for individual inputs. Such analyses can help vet model behaviour at an instance-level, though it is seldom obvious if the insights gained therein would generalise globally. For example, a local CE may suggest that a credit decisioning model is not biased against a protected attribute (e.g., gender, race), despite net biases existing across all inputs. A potential way to gain global insights is to aggregate local explanations, but given that the generation of CEs is generally computationally expensive, it is not evident that such an approach would scale well or even retain accuracy. % reference works that suggest this?

\citet{rawal2020individualized} investigates this problem, proposing Actionable Recourse Summaries (AReS), a framework that constructs global counterfactual explanations (GCEs). This work reports our attempt to understand and implement AReS. Although a useful and flexible framework, there exist shortcomings that limit its real-world use. Specifically, we find that AReS is a) computationally expensive and b) sensitive to continuous features, due to a dependency on the cardinality of the set used in the selection GCEs. We propose amendments to the algorithm and demonstrate that these lead to significant performance improvements on two benchmarked financial datasets.

\section{Investigations: Background, Motivation and Existing Methods}

\subsection{Local Counterfactual Explanations}
%Current counterfactual methods? Limitations/dangers of counterfactuals (current dilemmas) e.g., robustness, lack of optimality? Not that relevant, but can include.

%Saumitra
\cite{wachter2018counterfactual}~is one of the earliest works introducing CEs in the context of understanding black-box ML models. Their approach defines CEs as points that are close to the query input, w.r.t. some distance metric, that result in a desired model prediction. This work inspired several follow-up works where researchers proposed desirable properties of CEs and presented approaches to generate such CEs. For example,~\cite{mothilal2020explaining} argued that generating diverse CEs is essential for recourse. Other approaches aim to generate plausible CEs by considering proximity to the data manifold~\citep{poyiadzi2020face, looveren2021interpretable, kanamori2020dace} or by taking into account causal relations among input features~\citep{mahajan2019preserving}. Actionability of recourse is another important desideratum, as some features may be non-actionable, and hence should be excluded from the resulting CEs~\citep{ustun2019actionable}. In another direction, some works focused on generating CEs for specific model categories, such as tree-based models~\citep{lucic2021focus, tolomei2017interpretable, parmentier2021optimal}, or differentiable models~\citep{mothilal2020explaining}. For a detailed survey on CEs, please refer to~\citep{karimi2021survey, verma2020counterfactual}.

\subsection{Beyond Local Counterfactual Explanations: The Curse of Globality}
Despite a growing desire from practitioners for global explanation methods that provide summaries of model behaviour \citep{lakkaraju2022rethinking}, the struggles associated with summarising complex, high-dimensional models is yet to be comprehensively solved. Some manner of aggregations of local explanations has been suggested, though no compelling frameworks have been presented that a) are computationally tractable and b) return reliable GCEs. \citet{lakkaraju2022rethinking} also indicates a desire for increased interactivity with explanation tools, alongside global summaries, but these desiderata cannot be paired until the efficiency issues associated with global methods are addressed.

Such works have been few and far between. \citet{plumb2020explaining} and \citet{ley2021diverse} have sought global translations which transform each input point within a group to another desired target group, in the context of low-dimensional spaces. Meanwhile, \citet{becker2021global} provides an original method for GCE search, though openly struggles with scalability. To the best of our knowledge, only the aforementioned AReS specifically focuses on finding GCEs in the context of recourse.

\subsection{Beyond Individualized Recourse: Actionable Recourse Summaries (AReS)}
\label{sec:ares}

Recent work \citep{rawal2020individualized} proposes AReS, a comprehensive, model-agnostic framework for GCE generation. Building on the previously proposed two level decision sets \citep{lakkaraju2019faithful}, AReS adopts an original, interpretable structure, termed two level recourse sets.

\begin{figure}[h]
\centering
\includegraphics[width=\textwidth]{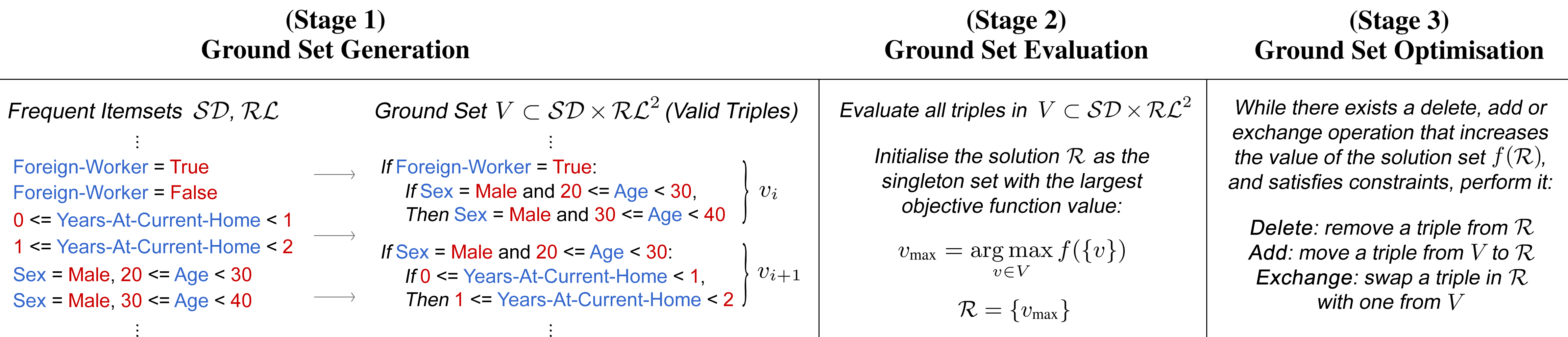}
\vspace{-0.2cm}
\caption{\small Workflow for our AReS implementation (without improvements). $\mathcal{SD}$ and $\mathcal{RL}$ are assigned to the same set generated by apriori. $\mathcal{SD}\times\mathcal{RL}^2$ is iterated over to compute all valid triples (Outer-If/Inner-If/Then conditions) for the ground set $V$ (Stage 1). Each item in $V$ is evaluated (Stage 2), and the optimisation procedure in \citet{lee2009nonmonotone} is applied (Stage 3), returning the smaller two level recourse set, $R$. A more detailed version of the framework can be found in both \citet{rawal2020individualized} and Appendix~\ref{app:implementation}.}
\label{fig:aresworkflow}
\end{figure}

A two level recourse set contains triples of the form Outer-If/Inner-If/Then conditions, pictured in Figure~\ref{fig:aresworkflow}. A frequent itemset mining algorithm such as apriori \citep{agrawal1994apriori} is deployed to generate candidate sets of conditions (e.g., Sex = Male, $20\leq$ Age $<30$). These are combined to generate triples, with all valid triples\footnote{A valid triple requires that the features in the Outer-If/Inner-If conditions do not match, and the features in the Inner-If/Then conditions match exactly with at least one change in feature value.} forming the ground set $V$. The candidate set of Outer-If conditions is termed $\mathcal{SD}$ (the subgroup descriptors), while $\mathcal{RL}$ denotes the candidate set used to select Inner-If or Then conditions. For apriori mining, the probability of an itemset in the data, or support threshold $p$, determines the size of $\mathcal{SD}$ and $\mathcal{RL}$, and consequently the size of $V$.

The subgroup descriptors $\mathcal{SD}$ can be set by the user to subgroups of interest, which is shown useful in assessing fairness via the disparate impact of recourses between subgroups. Otherwise, \citet{rawal2020individualized} assign $\mathcal{SD}$ and $\mathcal{RL}$ to the same set generated by apriori. AReS deploys a non-monotone submodular maximization algorithm \citep{lee2009nonmonotone} that selects, from the ground set $V$, a final, smaller set of rules $R$. Interpretability constraints for the total number of triples $\epsilon_1$, the maximum width of any Outer-If/Inner-If combination $\epsilon_2$ and the number of unique subgroup descriptors $\epsilon_3$ in $R$ are applied throughout. As in AReS, we take $\epsilon_1, \epsilon_2, \epsilon_3=20,7,10$.

While a novel framework, with an easily interpretable structure, AReS can fall short on two fronts:
\vspace{-0.2cm}

\paragraph{Computational Efficiency} An extremely low $p$ value is required to achieve high-performance, resulting in an impractically large ground set to optimise. Our work efficiently generates denser, higher-performing ground sets, unlocking the utility that practitioners have expressed desire for.
\vspace{-0.2cm}
\paragraph{Continuous Features} AReS proposes binning continuous features prior to generating frequent itemsets with apriori. However, we find that for models trained on continuous features, this approach struggles to trade speed with performance. Too few bins results in unrealistic recourses, but too many bins results in excessive computation time for apriori. We propose a modified ground set generation algorithm that demonstrates significant improvements on continuous data.

\pagebreak

\vspace{-0.4cm}
\section{Implementations: Actionable Recourse Summaries (AReS)}
\vspace{-0.3cm}

Our implementations include the original AReS framework, which follows the workflow demonstrated in Figure~\ref{fig:aresworkflow}, as well as optimisations. % adopting the conventions stated in \citet{rawal2020individualized} List these conventions? 
The ground set $V$ is defined as the set of triples from which the submodular maximisation algorithm \citep{lee2009nonmonotone} selects a two level recourse set $R\subset\mathcal{SD}\times\mathcal{RL}^2$, as stated in AReS.\footnote{Although \citet{rawal2020individualized} denote the solution to be a subset $R\subset\mathcal{SD}\times\mathcal{RL}$, this is mathematically impossible given that we require three conditions to form a valid triple (unless $\mathcal{RL}$ contains If/Then sets, which cannot be true if $\mathcal{SD}=\mathcal{RL}$, as AReS suggests). Correspondence with the authors confirms this.}
We denote the dataset as $\mathcal{X}$, and the set of affected individuals with an unfavourable prediction from the model as $\mathcal{X}_\text{aff}$.
The objective function $f(R)$ to be maximised is positive, comprising of incorrectness, coverage and cost.
The metrics used in evaluating performance are recourse accuracy (the percentage of instances in $\mathcal{X}_\text{aff}$ that are provided with a successful recourse), denoted $acc(R)$, and average recourse cost (the average cost of those individuals in $\mathcal{X}_\text{aff}$ for whom prescribed
recourses results in desired outcomes), denoted $cost(R)$. Owing to space constraints, we refer readers to \citet{rawal2020individualized} and Appendix~\ref{app:implementation} for full details.

The overall global counterfactual search in AReS for a two level recourse set, can be partitioned into three stages, as detailed in Figure~\ref{fig:aresworkflow} and Table~\ref{tab:aresimprovements}. We generate $V$, evaluate $V$, and optimise $V$ (selecting a smaller, more interpretable set, $R$). We describe each of these stages in detail below, alongside our respective optimisations. $R$ is evaluated in terms of recourse accuracy and average recourse cost, and it should be noted that, since recourse accuracy is monotonic (a new triple cannot invalidate a previous triple), $|R|\leq|V|\implies acc(R)\leq acc(V)$, providing us with an upper-bound.
\vspace{-0.4cm}

\subsection{Ground Set Generation (Stage 1)}
\label{sec:implementations1}
\vspace{-0.2cm}

The optimisation algorithm \citep{lee2009nonmonotone} requires a ground set $V$, which is generated by iterating through $\mathcal{SD}\times\mathcal{RL}^2$ and selecting valid triples. To generate larger $\mathcal{SD}$ or $\mathcal{RL}$, and thus larger $V$, a smaller apriori threshold $p$ is used. With no user input, we assign $\mathcal{SD}$ and $\mathcal{RL}$ to the same set generated by apriori, giving $V\subset\mathcal{RL}^3$, a strict subset.\footnote{We are guaranteed to find invalid triples in $\mathcal{RL}^3$. For example, if the first element of $\mathcal{RL}$ is ``Sex = Female'', the first iteration generates the triple ``If Sex = Female, If Sex = Female, Then Sex = Female'', an invalid triple.} We denote $|\mathcal{RL}|=n\implies|V|<n^3$.
Interpretability constraints that are independent of the optimisation, such as $\epsilon_2$, are applied in this stage in $\mathcal{O}(n^2)$ and not $\mathcal{O}(n^3)$ time (see Appendix~\ref{app:implementation1}). We introduce two methods to generate $V$. The first method computes an identical $V$ more efficiently, while the second computes a different $V$.
\vspace{-0.6cm}
\paragraph{Contribution 1a ($\bm{\mathcal{RL}}$-\textit{Reduction})} Iterating naively over $\mathcal{SD}\times\mathcal{RL}^2$ is wasteful, as many members of $\mathcal{RL}$ will never form valid ``If-Then'' conditions. We iterate instead over $\mathcal{RL}$ in $\mathcal{O}(n)$ time and compute feature combinations, before removing any items that contain a feature combination that only occurs once, yielding a new $\mathcal{RL}$ with size $\alpha n$, where $0\leq\alpha\leq1$ (note that $\mathcal{SD}=\mathcal{RL}$ is left untouched). For instance, the item ``Foreign-Worker = True, Sex = Male'' has a feature combination of ``Foreign-Worker, Sex''; if this only occurs once, it can be safely removed. For a given $\mathcal{RL}$, the ground set $V$ is the same as the original method, yet $(1-\alpha^2)n^3-n$ iterations are saved.%typically introducing speedups of around 10 times (see Experimental Results and Appendix X).
\vspace{-0.3cm}
\paragraph{Contribution 1b (\textit{Then-Generation}, $\pmb{q}$)} Instead of searching $\mathcal{SD}\times\mathcal{RL}^2$ for triples, we search $\mathcal{SD}\times\mathcal{RL}$ for If conditions, and deploy a separate method to generate Then conditions. Specifically,
\pagebreak

\begin{table}[h]
\begin{center}
\small
\begin{tabular}{c|c|c|c}
\multicolumn{1}{c}{} &\multicolumn{1}{c}{\bf (Stage 1)} &\multicolumn{1}{c}{\bf (Stage 2)} &\multicolumn{1}{c}{\bf (Stage 3)}\\
\multicolumn{1}{c}{} &\multicolumn{1}{c}{\bf Ground Set Generation} &\multicolumn{1}{c}{\bf Ground Set Evaluation} &\multicolumn{1}{c}{\bf Ground Set Optimisation}\\
\hline \\
\textbf{AReS} & $n^3$ Iterations Performed & Evaluates Full Ground Set & Searches Full Ground Set\\ &&&\\
\textbf{Ours} & $\alpha^2n^3+n$ or $n^2m$ & Evaluates and Shrinks Full & Searches Shrunk and\\
\ & Iterations Performed & or Partial Ground Set & Sorted Ground Set\\
\end{tabular}
\end{center}
\caption{\small A summary of our AReS enhancements w.r.t. each stage of the search. Definitions in Section~\ref{sec:implementations1}.}
\label{tab:aresimprovements}
% \begin{center}
% \begin{tabular}{c|c|c|c}
% \multicolumn{1}{c}{} & \multicolumn{1}{c}{\bf Stage of GCE Search} &\multicolumn{1}{c}{\bf AReS} &\multicolumn{1}{c}{\bf Ours}\\
% \hline \\
% 1 & Ground Set Generation & Performs $|\mathcal{RL}|^3$ Iterations & $\alpha^2|\mathcal{RL}|^3$ or $|\mathcal{RL}|^2|\mathcal{RL}_2|$\\ 
% 2 & Ground Set Evaluation & Evaluates Full Ground Set & Evaluates Full/Partial Ground Set\\
% 3 & Ground Set Optimisation & Searches Full Ground Set & Searches Reduced Ground Set\\
% \end{tabular}
% \end{center}
% \caption{\small AReS Enhancements (alternative format).}
\end{table}
\vspace{-0.1cm}
for each valid element of $\mathcal{SD}\times\mathcal{RL}$, with index $i$, we compute its feature combination and filter the dataset by these features (also removing inputs that satisfy the initial If conditions), before applying apriori again, with threshold $q$, to generate a set of Then conditions, denoted $\mathcal{T}_i$. We can lower bound $q$ as $1/|\mathcal{X}|$ (no observed itemset can have frequency $<1$), and we find that varying $q$ has little impact on speed but reduces performance (Appendix~\ref{app:experiments}). If $m=\text{max}_i|\mathcal{T}_i|$ is the maximum size of any such $\mathcal{T}_i$, the number of iterations has an upper bound of $n^2m$.  The ground set generated differs from the original method and we observe significant improvements on continuous features.

\subsection{Ground Set Evaluation (Stage 2)}
\vspace{-0.1cm}

The submodular maximisation \citep{lee2009nonmonotone} first evaluates the objective function $f$ over all triples $v\in V$, before initialising the solution $R$ as the singleton set $\{v\}$ with the maximum $f(\{v\})$. For large $|V|$, this evaluation becomes computationally costly (more-so does the subsequent ground set optimisation), and many triples are also redundant. However, we require large $|V|$ in order to find high-performing triples and achieve an acceptable upper bound\footnote{For instance, if $acc(V)=25$\%, we cannot achieve $acc(R)>25$\%; conversely, a ground set with $acc(V)=80$\% requires major evaluation and will also include many low-performing, redundant triples.} on the final set, $R\subseteq V$.

\begin{wrapfigure}[11]{r}{0.305\textwidth}
\vspace{-0.45cm}
\centering
\includegraphics[width=0.295\textwidth]{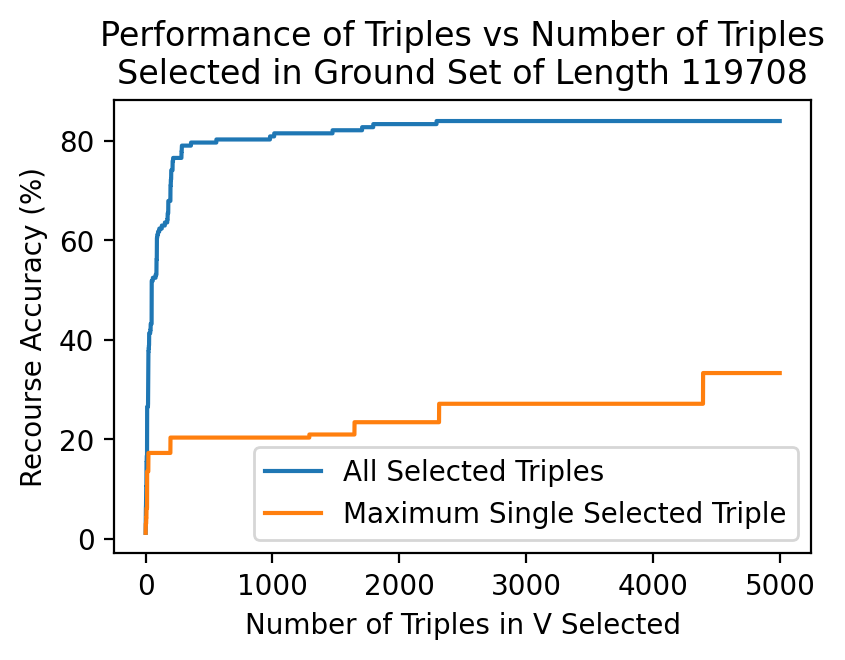}
\vspace{-0.3cm}
\caption{\small Redundancy in ground set $V$. German Credit, $p=0.22$.}
\label{fig:accuracyvsgroundset}
\end{wrapfigure}
%The objective function $f$ takes into account recourse correctness, coverage and cost, as defined in \citet{rawal2020individualized}.
We take advantage of two empirical observations: the generation of a large ground set $V$ is relatively cheap; and the recourse accuracy $acc(V)$ of the full ground set is approached far before the whole set has been evaluated. This allows us to efficiently shrink large ground sets to smaller ones with comparable recourse accuracy. For example, in 40 seconds, the apriori threshold $p=0.22$ on the German Credit dataset produces a ground set with $|V|=119708$. While $acc(V)=84$\% then takes 300 seconds to evaluate, 84\% is converged to after only 5 seconds (Figure~\ref{fig:accuracyvsgroundset}). The maximum value of a single triple is also seen to converge quickly. We can generate a large ground set, before only evaluating a small portion of this set to yield an equally high-performing yet denser ground set. Note that simply raising $p$ to 0.323 and producing a smaller ground set of equal size does not yield 84\% accuracy (instead, it yields 27\%).
\vspace{-0.2cm}
\paragraph{Contribution 2 (\textit{V-Reduction}, $\pmb{r}, \pmb{r'}$)} We evaluate a fixed number of triples and form a new ground set in one of two ways: by adding each new triple, or by only adding triples that increase the recourse accuracy of the new ground set (i.e. vertical steps in Figure~\ref{fig:accuracyvsgroundset}, blue). We denote these $r$ and $r'$ respectively. For example, $r'=$ 2000 results in 2000 evaluations and less than 2000 triples added.

\subsection{Ground Set Optimisation (Stage 3)}

The bottleneck in the AReS framework is, however, the submodular maximisation in \citet{lee2009nonmonotone}, which takes the ground set $V$ and returns a reduced set $R$ that satisfies the interpretability constraints. The time taken is a function of the size $|V|$ of the ground set; we can thus achieve speedups by effectively further shrinking the ground set pre-optimisation. The submodular maximisation provides optimality guarantees. As such, we do not modify the algorithm itself.\footnote{Importantly, however, with knowledge of our upper bound $acc(R)\leq acc(V)$, optimisation can be terminated if this bound is approached. Such a bound can also be used to determine if Stage 3 is even initiated.} Our ground set modifications instead provide the algorithm with a superior starting point and upper bound.
\vspace{-0.2cm}
\paragraph{Contribution 3 (\textit{V-Selection}, $\pmb{s}$)} We propose to sort the (new) ground set by recourse accuracy (already computed), and select the $s$ highest-performing triples. If $s=r$ or $r'$, no sorting occurs.

\pagebreak

\section{Improvements: Experimental Results}
\label{sec:experiments}

We evaluate our methods on two benchmarked financial datasets: the German Credit dataset \citep{dua2019uci} classifies credit risk on people described by a set of attributes, consisting mostly of categorical features; the HELOC (Home Equity Line of Credit) dataset \citep{fico2018heloc} includes anonymised credit applications made by real homeowners, and consists solely of continuous features. We train Deep Neural Networks (DNNs) with width 50 and depth 10 and 5 respectively on these datasets, with an 80\% training split. Continuous features are binned into 10 equal intervals post-training (see Section~\ref{sec:ares} trade-off), and recourses are constructed on the training set.

We analyse the performance of AReS and our improvements cumulatively, at each stage of the workflow. For various input parameter combinations ($p$, $r$, $r'$ and $s$), the final two level recourse sets returned in Stage 3 achieve significantly higher recourse accuracy within a time frame of 300 seconds (5 minutes), achieving accuracies for which AReS required 45 minutes on German Credit, and over 18 hours on HELOC. Further hyper-parameter details are located in Appendix~\ref{app:experiments}.

\begin{figure}[h]
\centering
\includegraphics[width=\textwidth]{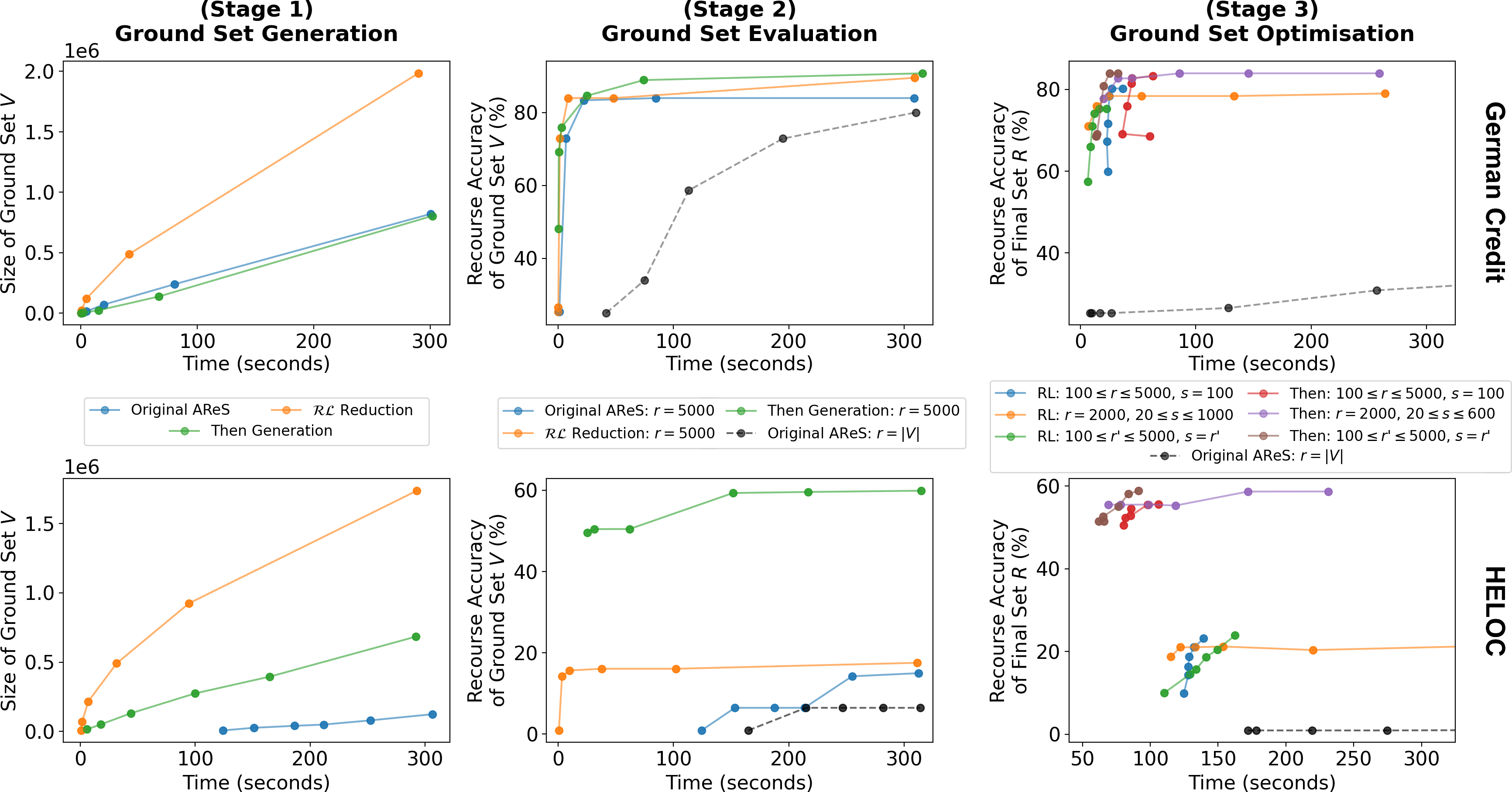}
\vspace{-0.4cm}
\caption{\small Computational Improvements. Top Row: German Credit. Bottom Row: HELOC. Left: Size of Ground Set $V$ vs Time. Centre: Ground Set $acc(V)$ vs Time. Right: Final Set $acc(R)$ vs Time.}
\label{fig:allexperiments}
\end{figure}\vspace{-0.1cm}
\paragraph{Takeaways} In Stage 1, we demonstrate that $\mathcal{RL}$-\textit{Reduction} is capable of generating an equivalent ground set $V$ orders of magnitude faster than the original method. Our \textit{Then-Generation} technique also constructs (different) ground sets rapidly. Stage 2 \textit{V-Reduction} ($r=5000$) performs significantly better than full evaluation, and \textit{Then-Generation} erases many of the limitations surrounding continuous features. We finally observe vast speedups in Stage 3, owing to the construction of small yet high-performing ground sets: $r$, $r'$ and $s$ restrict the size of $V$ yet retain a near-optimal $acc(V)$.

We note that the choice of $\gSD=\gRL$ affects performance (selecting a fixed $\gSD$ would reduce the size of $|\gX_\text{aff}|$ and $V$) though argue then that we are emphasising the scalability of our new approach.

\vspace{-0.1cm}
\section{Conclusion}
\vspace{-0.1cm}
This work studies the current state of global counterfactual explanations (GCEs), and addresses in detail the scalability/performance issues we find in the recently proposed AReS framework \citep{rawal2020individualized}. We investigate works on both global and local counterfactual explanations before implementing and improving AReS. With mounting desire from a practitioner viewpoint for access to fast, interactive explainability tools \citep{lakkaraju2022rethinking}, it is crucial that such methods are not inefficient. We propose improvements to the AReS framework that speed up the generation of GCEs by orders of magnitude, also witnessing significant accuracy improvements on continuous data. Our hope is that this will inspire further research into the particularly under-studied area of GCEs, and prove useful as the development of explainability tools grows in the coming years.

\pagebreak

\paragraph{Acknowledgments} We thank the original authors Kaivalya Rawal and Himabindu Lakkaraju for their helpful discussion of the proposed AReS framework in \cite{rawal2020individualized}.

\paragraph{Disclaimer}
This paper was prepared for informational purposes by
the Artificial Intelligence Research group of JPMorgan Chase \& Co. and its affiliates (``JP Morgan''),
and is not a product of the Research Department of JP Morgan.
JP Morgan makes no representation and warranty whatsoever and disclaims all liability,
for the completeness, accuracy or reliability of the information contained herein.
This document is not intended as investment research or investment advice, or a recommendation,
offer or solicitation for the purchase or sale of any security, financial instrument, financial product or service,
or to be used in any way for evaluating the merits of participating in any transaction,
and shall not constitute a solicitation under any jurisdiction or to any person,
if such solicitation under such jurisdiction or to such person would be unlawful.

% Acknowledge Hima/Kai

\bibliography{ref}
\bibliographystyle{iclr2022_conference}

\pagebreak
\appendix
\section*{Appendix}
This appendix is formatted as follows.
\begin{enumerate}
    \item We discuss the \textit{Datasets and Models} used in our work in Appendix~\ref{app:datasets}.
    \item We discuss the \textit{Implementation Details} of our work in Appendix~\ref{app:implementation}.
    \item We list the \textit{Experimental Details} of our work and analyse \textit{Further Results} in Appendix~\ref{app:experiments}.
\end{enumerate}

\section{Datasets and Models}
\label{app:datasets}

Two benchmarked financial datasets are employed in our experiments, both of which are a) binary classification and b) publicly available. Details are provided below and in Table~\ref{tab:datasets}. Our experiments include just one type of model, Deep Neural Networks, which we also describe below and in Table~\ref{tab:models}.

\subsection{Datasets}

The \textbf{German Credit} dataset \citep{dua2019uci} can be obtained from and is described in detail at the following URL: {\footnotesize\url{https://archive.ics.uci.edu/ml/datasets/statlog+(german+credit+data)}}. We augment input dimensions by performing a one-hot encoding over necessary variables (Sex, Foreign-Worker, etc). The documentation for this dataset also details a cost matrix, where false positive predictions induce a higher cost than false negative predictions, but we ignore this in model training. Note that this is distinct from the also common Default Credit dataset.

The \textbf{HELOC} (Home Equity Line of Credit) dataset \citep{fico2018heloc} can upon request be obtained from and is described in detail at the following URL: {\footnotesize\url{https://community.fico.com/s/explainable-machine-learning-challenge}}. Missing values in the dataset are represented with negative integers; we drop inputs where all feature values are missing, and replace the remaining missing values in the dataset with the median value for that feature. We also drop any duplicate inputs in the dataset. Notably, the majority of features are monotonically increasing/decreasing.

\begin{table*}[hb]
\centering
\begin{tabular}{|c|c|c|c|c|c|c|}
\bottomrule
Name & Categorical & Continuous & Input Dim. & No. Train & No. Test\\
\toprule
\bottomrule
German Credit & 17 & 3 & 71* & 800 & 200\\
\hline
HELOC & 0 & 23 & 23 & 7896* & 1975*\\
\toprule
\end{tabular}

\vspace{0.1cm}
\small *Denotes values post-processing (one-hot encoding inputs, dropping inputs).
\caption{\small Summary of the datasets used in our experiments. Although German Credit includes continuous features, we find that they have limited effect on the model both during training and in the resulting explanations.}
\label{tab:datasets}
\end{table*}

\subsection{Models}

We train Deep Neural Networks (DNNs) with width 50 and depth 10 and 5 respectively on these datasets, with an 80\% to 20\% train to test split. Layers include dropout, bias and ReLU activation functions. We map the final layer to the output using softmax, and use Adam to optimise a cross-entropy loss function in the standard manner. Table~\ref{tab:models} details various model parameters/behaviours.

\begin{table*}[ht]
\centering
\begin{tabular}{|c|c|c|c|c|c|c|c|}
\bottomrule
Name & Width & Depth & Dropout & Train Acc. & Test Acc & $|\gX_\text{aff}|$ & $|\gX_\text{aff}|/|\gX|$\\
\toprule
\bottomrule
German Credit & 50 & 10 & 0.3 & 82\% & 79\% & 162 & 20\%\\
\hline
HELOC & 50 & 5 & 0.5 & 74\% & 73\% & 3882 & 49\%\\
\toprule
\end{tabular}
\caption{\small Summary of the DNNs used in our experiments. The proportion of negative labels in the dataset were 30\% and 53\% for German Credit and HELOC respectively; our models roughly follow suit (20\% and 49\%).}
\label{tab:models}
\end{table*}

Of note is the scalability of AReS, which struggled with HELOC, a dataset that contained significantly more points to explain ($|\gX_\text{aff}|$) than German Credit. Additionally, the proportion of points with positive predictions (80\% for German Credit and 51\% for HELOC) influences the ease with which AReS finds recourses. For stringent models (those which scarcely predict positively), it would make sense that the vast majority of frequent itemsets generated by apriori are representative of feature value combinations that exist in the inputs with negative predictions, and we might therefore expect to need to generate an enormous number of triples before we can identify successful recourses.

\section{Implementation Details}
\label{app:implementation}

We use this Appendix to provide further details regarding the implementation of each stage of the AReS workflow. Our implementation of AReS, without improvements, does in fact differ slightly from that proposed in \cite{rawal2020individualized}, and as such we will justify our changes herein. We of course acknowledge that this implementation is far from the most efficient possible, though hope that the patterns and improvements we have identified can aid further development of not only this framework, but others in the global counterfactual explanations space also.

\subsection{Ground Set Generation (Stage 1)}
\label{app:implementation1}

As stated in the main text, our implementation applies constraints during ground set generation where possible. AReS includes interpretability constraints for the total number of triples $\epsilon_1$, the maximum width of any Outer-If/Inner-If combination $\epsilon_2$ and the number of unique subgroup descriptors $\epsilon_3$ in $R$. As in AReS, we take $\epsilon_1, \epsilon_2, \epsilon_3=20,7,10$. In our implementation, we expedite the $\epsilon_2$ width constraint to the ground set generation process by constraining apriori to only return frequent itemsets that have length $\epsilon_2-1$ or less, since those already with width $\epsilon_2$ cannot then be further combined with another itemset to form Outer-If/Inner-If conditions. If the width constraint is not violated for the If conditions, the resulting triple will automatically satisfy the constraint.

The implication of this is that we can apply the constraint in Stage 1 while we generate the ground set (in the first two levels of the iteration through $\gRL^3$). This avoids applying the width constraint mid-optimisation in Stage 3, reducing the time complexity of the operation from $\gO(n^3)$ to $\gO(n^2)$. It also reduces the number of constraints used in \citet{lee2009nonmonotone}, speeding up Stage 3. Since it makes sense that triples which violate the maximum width condition should not be generated in Stage 1, we assume that a similar approach is deployed (though not stated) in \citet{rawal2020individualized}.

\paragraph{Then-Generation} A lower bound for the threshold $q$ used in Then-Generation was also alluded to in the main text. In fact, there always exists a lower bound when mining frequent itemsets, such as in apriori, since no observed itemset can occur less than once. Thus, setting $q<1/|\gX|$ would be redundant. This allows us to analyse the full effect of $1/|\gX\leq q\leq 1$ in Appendix~\ref{app:experiments}.

\subsection{Ground Set Evaluation (Stage 2)}
\label{app:implementation2}

Our improvement (Contribution 2) evaluates the objective function $f$ (see Section~\ref{app:implementation3}) over a fixed number of triples in $V$ (recall that AReS evaluates the entirety of $V$). As we've demonstrated empirically, albeit on the two datasets tried in this investigation, evaluating the entire ground set is wasteful, given that performance of the first $r$ elements of $V$ saturates quickly, and more so if one considers that Stage 3 must then perform submodular maximisation over a space potentially hundreds of times as large, and that \citet{lee2009nonmonotone} only guarantees polynomial time.

However, there is a distinction between evaluating the objective function $f$ and evaluating the $acc$ and $cost$ terms used in evaluation. Fortunately, no extra major computation is required to evaluate the $acc$ and $cost$ terms, since the objective function $f$ returns model predictions and costs, and although the two processes differ, they can be carried out efficiently in tandem. This is promising, as not only does our method allow us to terminate evaluation once saturation has been reached, but it also provides us with the upper bound $acc(R)\leq acc(V)$. In many of our experiments, this upper bound is actually reached in Stage 3 far before the algorithm has finished, presenting us with a straightforward opportunity for early termination of the algorithm. This could further save time dramatically, though was not included in our experiments.

\subsection{Ground Set Optimisation (Stage 3)}
\label{app:implementation3}

We introduce two key modifications to Stage 3 of our implementation. The first is to the objective function, the second is to the submodular maximisation in \cite{lee2009nonmonotone}.

\paragraph{Objective Function} The objective function $f(R)$ in \citet{rawal2020individualized} is designed to be non-normal, non-negative, non-monotone and submodular, and to have constraints that are matroids. These conditions are required for the submodular maximisation in \citet{lee2009nonmonotone} to have a formal guarantee of convergence. This results in four terms in $f(R)$: \textit{incorrectrecourse}, \textit{cover}, \textit{featurecost}, \textit{featurechange}. Bar the \textit{cover} term, all of these are subtracted from $f(R)$ (i.e., maximising correct recourse by maximising the negative of \textit{incorrectrecourse}). Such an objective function with three adjustable hyperparameters can be very difficult to tune. For that reason, we also trial in our experiments an objective that consists very simply of $acc(R)-\lambda\times cost(R)$, which we maximise. We argue that the formal guarantees of convergence (polynomial time) are largely a misdirection of efforts in the original method. Polynomial time is not particularly helpful when the size of ground sets required for certain datasets/models is huge, and thus we instead focus on reducing the size of the ground set while retaining quality before the submodular maximisation \citep{lee2009nonmonotone} is applied.

\paragraph{Submodular Maximisation} The algorithm states that, for $k$ constraints, you can exchange up to $k$ elements from your solution set $R$ alongside the addition of one element from $V$. Stated also is that the optimisation should be repeated $k+1$ times, before the best solution for $R$ is then chosen. In reality, both of these induce high computational costs. Trivially, for the latter, ignoring the maximum width constraint (Appendix~\ref{app:implementation1}) and taking $k+1=3$, we will mostly increase the time taken by AReS three-fold. Having observed that both of these steps do not improve the performance of AReS significantly in our experiments, we omit them from the original and improved implementations.

\section{Experimental Details and Further Results}
\label{app:experiments}

We use the training data from each dataset to learn recourses in our experiments (future work could analyse the effectiveness of such rules on unseen test data). Since AReS struggles to achieve sufficient recourse accuracy within reasonable time-frames for our datasets and models, we set the hyperparameters for \textit{featurecost} and \textit{featurechange}, or $\lambda$, to 0, also finding that the average cost of recourses were low and did not vary a large amount, justifying the decision to target correctness. The remaining hyperparameters used in the Figure~\ref{fig:allexperiments} experiments (Section~\ref{sec:experiments}) are as detailed in Table~\ref{tab:experiments}.
\begin{table}[b]
\scriptsize
\centering
\begin{tabular}{p{0.08\textwidth}|p{0.24\textwidth}|p{0.26\textwidth}|p{0.28\textwidth}}
 & {\small\hspace{0.09\textwidth}Stage 1} & {\small\hspace{0.09\textwidth}Stage 2} & {\small\hspace{0.09\textwidth}Stage 3}\\
\hline
\ 

{\small German 

\ Credit} &
\textbf{OG}: $0.169\leq p\leq0.390\ \longrightarrow$

\textbf{RL}: \ \ \ $0.39\leq p\leq0.149\ \longrightarrow$

\textbf{Then}: \ $0.9\leq p\leq0.303\ \longrightarrow$

\ \ \ \ \ \ \ \ \ \ \ \ $q=0.00125$&

\textbf{OG}: \ \ \ $r=5000$

\textbf{RL}: \ \ \ $r=5000$ 

\textbf{Then}: $r=5000, q=0.00125$

\textbf{OG}: $0.316\leq p\leq0.26$, $r=|V|$ &

\textbf{OG}: \ \ $0.39\leq p\leq0.305$, $r=|V|$

\textbf{RL}: \ \ \ $p=0.245$ 

\textbf{Then}: $p=0.48$,

\ \ \ \ \ \ \ \ \ \ \ $q=0.00125$\\
\hline
\ 

{\small HELOC}&
\textbf{OG}: \ $0.325\leq p\leq0.285\longrightarrow$

\textbf{RL}: \ \ $0.325\leq p\leq0.203\longrightarrow$

\textbf{Then}: $0.75\leq p\leq0.563\longrightarrow$

\ \ \ \ \ \ \ \ \ \ \ $q=0.000127$&

\textbf{OG}: \ \ \ $r=5000$

\textbf{RL}: \ \ \ $r=5000$ 

\textbf{Then}: $r=5000, q=0.000127$

\textbf{OG}: $0.325\leq p\leq0.3$, $r=|V|$ &

\textbf{OG}: \ \ $0.324\leq p\leq0.318$, $r=|V|$

\textbf{RL}: \ \ \ $p=0.245$

\textbf{Then}: $p=0.48$,

\ \ \ \ \ \ \ \ \ \ \ $q=0.000127$\\
\end{tabular}
\caption{\small The keys \textbf{OG} (\textit{Original AReS}), \textbf{RL} ($\gRL$-\textit{Reduction}) and \textbf{Then} (\textit{Then-Generation)} refer to the generation process of the ground set, as per Section~\ref{sec:implementations1}. Arrows indicate values carried from one stage to the next. Apriori thresholds $p$ and $q$ are listed. Remaining parameters $r$, $r'$ and $s$ are listed in the original Figure~\ref{fig:allexperiments} plots.}
\label{tab:experiments}
\end{table}

\begin{wrapfigure}[18]{r}{0.5\textwidth}
\vspace{-0.45cm}
\centering
\includegraphics[width=0.49\textwidth]{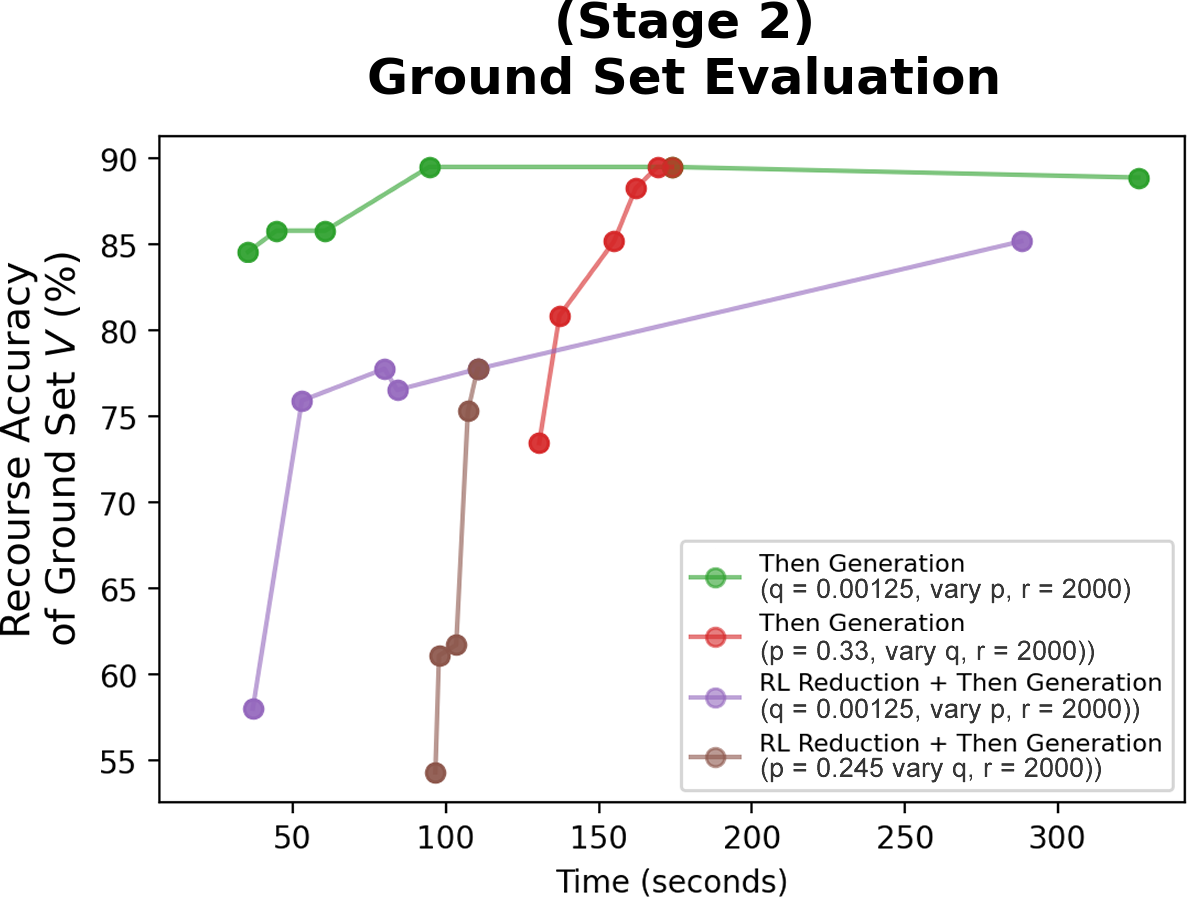}
\vspace{-0.2cm}
\caption{\small Effect of apriori threshold $q$ in the proposed \textit{Then-Generation} method (German Credit).}
\label{fig:effectofq}
\end{wrapfigure}
Recall also that we have bounded the range of the apriori threshold $q$ used in \textit{Then-Generation} to $1/|\gX|\leq q\leq 1$ (Section~\ref{sec:implementations1} and Appendix~\ref{app:implementation1}). Figure~\ref{fig:effectofq} demonstrates that for $q>1/|\gX|$, we slightly reduce the time taken by the algorithm, at the expense of a much larger drop in performance. Observe that the red and brown lines (where $p$ is held constant and $q$ is varied) converge to the green and purple lines (where $q=1/|\gX|$ and $p$ is varied) respectively. The brown and purple plots also indicate that combining our two improvements $\gRL$-\textit{Reduction} and \textit{Then-Generation} performs sub-optimally. We thus decide to evaluate these improvements separately with a fixed $q=1/|\gX|$ threshold used in the \textit{Then-Generation} method.

\end{document}